\documentclass[conference, 9pt]{IEEEtran}
\IEEEoverridecommandlockouts

\usepackage{cite}
\usepackage{amsmath,amssymb,amsfonts}
\usepackage{amsthm}
\usepackage{amssymb}
\usepackage{amsfonts}
\usepackage{booktabs}
\usepackage{multirow}
\usepackage{algorithmic}
\usepackage{graphicx}
\usepackage{textcomp}
\usepackage[table,xcdraw]{xcolor}
\usepackage{url}
\usepackage{makecell}
\usepackage{enumitem}
\def\BibTeX{{\rm B\kern-.05em{\sc i\kern-.025em b}\kern-.08em
    T\kern-.1667em\lower.7ex\hbox{E}\kern-.125emX}}
\begin{document}

\title{VisTa: Visual-contextual and Text-augmented\\ Zero-shot Object-level OOD Detection
\thanks{Research is supported by the Key Research and Development Program of Guangdong Province (grant No. 2021B0101400003).}
}

\author{\IEEEauthorblockN{
    Bin Zhang\IEEEauthorrefmark{1}\IEEEauthorrefmark{2},
    Xiaoyang Qu\IEEEauthorrefmark{2}\IEEEauthorrefmark{3},
    Guokuan Li\IEEEauthorrefmark{1}\IEEEauthorrefmark{3},
    Jiguang Wan\IEEEauthorrefmark{1} 
    and Jianzong Wang\IEEEauthorrefmark{2}}
    \IEEEauthorblockA{\IEEEauthorrefmark{1}Wuhan National Laboratory for Optoelectronics, Huazhong University of Science and Technology, Wuhan, China}
    \IEEEauthorblockA{\IEEEauthorrefmark{2}Ping An Technology (Shenzhen) Co., Ltd, Shenzhen, China}
    \IEEEauthorblockA{\{binz2398, quxiaoy\}@gmail.com, \{liguokuan, jgwan\}@hust.edu.cn, jzwang@188.com}
    \thanks{\IEEEauthorrefmark{3}Corresponding authors.}    
}

\maketitle

\begin{abstract}
As object detectors are increasingly deployed as black-box cloud services or pre-trained models with restricted access to the original training data, the challenge of zero-shot object-level out-of-distribution (OOD) detection arises. This task becomes crucial in ensuring the reliability of detectors in open-world settings. While existing methods have demonstrated success in image-level OOD detection using pre-trained vision-language models like CLIP, directly applying such models to object-level OOD detection presents challenges due to the loss of contextual information and reliance on image-level alignment. To tackle these challenges, we introduce a new method that leverages visual prompts and text-augmented in-distribution (ID) space construction to adapt CLIP for zero-shot object-level OOD detection. Our method preserves critical contextual information and improves the ability to differentiate between ID and OOD objects, achieving competitive performance across different benchmarks.
\end{abstract}


\section{Introduction}

As object detection models are being more frequently used in practical applications \cite{wosner2021object, bergmann2019mvtec, roy2023wildect}, ensuring their robustness in open-world scenarios is crucial. In such environments, models frequently encounter inputs not belonging to the training distribution, referred to as out-of-distribution (OOD) samples. In contrast, in-distribution (ID) samples refer to those on which the model has been trained. When deep neural networks encounter OOD samples, they often fail silently, leading to overconfident erroneous predictions\cite{miller2018dropout, dhamija2020overlooked, rosenfeld2018elephant, hendrycks2016baseline, bendale2016towards, guo2017calibration}. This failure to correctly identify OOD samples can have severe consequences, such as misclassification and overconfidence in predictions. In safety-critical tasks like autonomous driving, where undetected OOD objects can cause accidents, these consequences can be particularly risky\cite{automot_OOD, amodei2016concrete, sunderhauf2018limits}. Therefore, the emergence of object-level OOD detection, which focuses on identifying anomalous objects at a granular level, is a crucial and rapidly evolving research area.

In object-level OOD detection, prior works \cite{Siren, VOS, wu2023tib} often require integrating additional modules into the training process of detectors, leveraging the model's inherent uncertainty. These approaches typically necessitate supervised training and the inclusion of extra components, such as uncertainty estimation heads, to identify potential OOD objects. Additionally, the SAFE method \cite{wilson2023safe} avoids retraining the object detector. Instead, SAFE manullay extracts features from the pre-trained object detector and trains a separate MLP to identify OOD objects, reducing the need to modify the original detection pipeline and offering a more efficient and flexible alternative. Estimation-based methods \cite{ma_distance, wang2022vim, sun2022out_knn, tack2020csi}, on the other hand, focus on identifying outliers through techniques like distance metrics or contrastive learning, directly learning from the training data.

Despite their effectiveness, these methods are constrained by their dependence on ID data and often require retraining or adding new training phases. Furthermore, many pre-trained models, like those offered by Hugging Face Model Hub, are trained on proprietary or large-scale datasets that are not publicly available. This limitation makes it difficult for practitioners to fully leverage these models with traditional OOD detection techniques, as they need access to the original training data.

\begin{figure}
    \centering
    \includegraphics[width=1\linewidth]{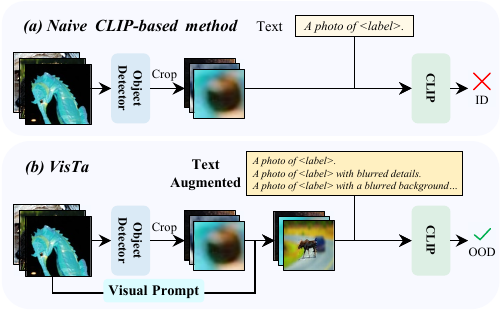}
    \caption{Framework of (a) the naive CLIP-based method and (b) the proposed VisTa approach. In (a), zero-shot CLIP-based OOD detection is directly adapted for object-level OOD detection using cropping. In (b), we emphasize the two key components of our VisTa method.}
    \label{fig:frame_compare}
    \vspace{-1em}
\end{figure}

To address these issues and make OOD detection more accessible, the zero-shot detection paradigm is a promising solution. Utilizing publicly available pre-trained models eliminates the need for ID-specific training data, providing a practical alternative for those with limited resources. Recently, the advent of pre-trained vision-language models (VLMs), such as CLIP\cite{CLIP}, ALIGN\cite{ALIGN}, BLIP\cite{li2022blip}, and InternVL\cite{chen2024internvl}, has opened new avenues for OOD detection. Previous work\cite{MCM, esmaeilpour2022zero, wang2023clipn} have focused on zero-shot OOD detection within image classification, mainly using CLIP as a general classifier. This study concentrates on zero-shot object-level OOD detection. As shown in Fig. \ref{fig:frame_compare}(a), while the performance on image-level tasks is excellent, directly applying it in object-level tasks with the help of techniques like cropping has a poor performance due to significant loss of contextual information.

\begin{figure*}
    \centering
    \includegraphics{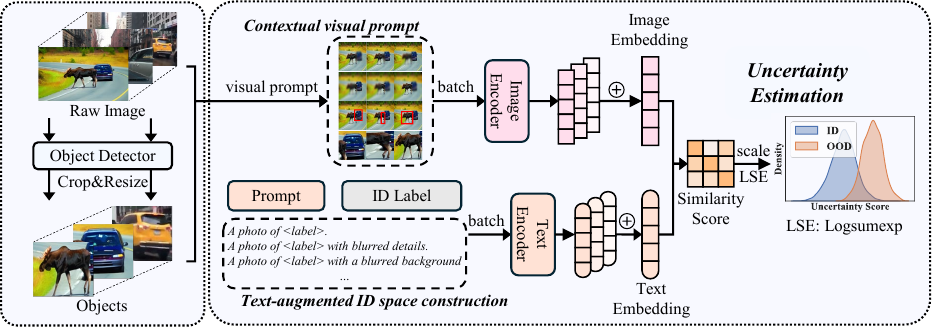}
    \caption{\textbf{Overview of VisTa.} The ID embedding space $\mathcal{C}$ is built with CLIP’s text encoder and augmented prompts. Image embeddings from contextual visual prompts are compared to $\mathcal{C}$ to compute similarity scores, which are used to calculate uncertainty and distinguish between ID and OOD samples.}
    \label{fig:overview}
\end{figure*}

Given these challenges, as shown in Fig. \ref{fig:frame_compare}(b), our proposed approach offers a novel zero-shot solution that leverages visual prompts to adapt CLIP for object-level OOD detection, eliminating the need for retraining or additional modules. Our prompting mechanism guides the model to emphasizes features at the object level while preserving essential contextual information lost during cropping, allowing CLIP to transfer its pre-trained knowledge more effectively. Additionally, we introduce a text-augmented strategy to enhance the construction of the ID embedding space, aligning textual information with the visual prompts to expand the representative capacity of the ID space. This enriched embedding space enables more accurate differentiation between ID and OOD objects, significantly improving detection performance.

Our contributions are as follows:
\begin{itemize}[nosep]
    \item We propose a visual prompt-based method that effectively adapts CLIP for object-level OOD detection, preserving crucial contextual information and capturing features more accurately.
    \item We introduce a text-augmented approach that enhances the ID embedding space, aligning it with the visual prompts to improve the model's ability to differentiate between ID and OOD objects.
    \item We conduct extensive experiments demonstrating that our zero-shot approach achieves strong performance, surpassing existing methods on multiple object-level OOD detection benchmarks.
\end{itemize}

\section{Method}

\subsection{Overview}

This work introduces a novel approach to enhance object-level OOD detection by adapting the CLIP model, which is traditionally used for image-level tasks. The key challenge in applying CLIP to object-level tasks is the loss of contextual information when focusing on individual objects. We propose two main components to address this: context-aware visual prompts and text-augmented ID space construction. The visual prompts guide CLIP to preserve contextual information better, improving its ability to detect OOD samples at the object level. Additionally, the text-augmented approach enhances the construction of the ID embedding space, expanding its representational capacity to align with the applied visual prompts. Together, these components enable a robust, zero-shot approach to object-level OOD detection, leveraging the strengths of multimodal representations and providing a reliable solution to the challenge of OOD detection.

The pipeline of our proposed approach is shown in Fig. \ref{fig:overview} and can be outlined as follows: (1) We first construct the ID embedding space $\mathcal{C}$ by encoding the $K$ ID class labels $\mathcal{Y}_\text{in}=\{y_i,\ i=1,\,2,\,3,\,...,\, K\}$ using CLIP’s text encoder, enhanced with augmented prompts to improve its expressiveness; (2) Next, consider an image $x$ along with its corresponding bounding box $b$, we generate image embeddings using contextual visual prompts, ensuring that crucial contextual information is preserved; (3) Finally, these image embeddings are compared with the text-augmented ID embedding space to compute the similarity score, which is then used to calculate the final uncertainty score, distinguishing between ID and OOD samples:
\begin{equation}
    G(x, b) = \begin{cases}
\text{in}, &\text{if}\  \mathbb{E}[\sigma(x, b) \mid \mathcal{C}] \leq \gamma\\
\text{out}, &\text{if}\  \mathbb{E}[\sigma(x, b) \mid \mathcal{C}] > \gamma\\
\end{cases}
\label{equ:case}
\end{equation}
where $\sigma(x,b)$ represents the extracted feature, $\mathbb{E}[\sigma(x, b) \mid \mathcal{C}]$ denotes the uncertainty score, and $G(x, b)$ is the predicted outcome for OOD detection. The threshold $\gamma$ is determined based on the distribution of ID data, ensuring that the majority (e.g., 95\%) of the ID data can be correctly distinguished from OOD data.




\begin{table*}[htbp]
\centering
\caption{\textbf{Comparison with other competitive OOD detection methods.} The comparison metrics are FPR95 and AUROC, where $\uparrow$ and $\downarrow$ denote preferred directions. All results are presented in percentages, with \textbf{bold} numbers indicating superior results.}
\label{tab:main_tab}
\begin{tabular}{@{}cccccccccc@{}}
\toprule
\textbf{ID dataset}                     & \multicolumn{4}{c}{\textbf{BDD-100K}} & \multicolumn{1}{l}{} & \multicolumn{4}{c}{\textbf{PASCAL-VOC}} \\ \midrule
\textbf{OOD dataset} &
  \multicolumn{2}{c}{\textbf{OpenImages}} &
  \multicolumn{2}{c}{\textbf{MSCOCO}} &
   &
  \multicolumn{2}{c}{\textbf{OpenImages}} &
  \multicolumn{2}{c}{\textbf{MSCOCO}} \\ \midrule
\textbf{Detection method} &
  \textbf{FPR95 $\downarrow$} &
  \textbf{AUROC $\uparrow$} &
  \textbf{FPR95 $\downarrow$} &
  \textbf{AUROC $\uparrow$} &
  \textbf{} &
  \textbf{FPR95 $\downarrow$} &
  \textbf{AUROC $\uparrow$} &
  \textbf{FPR95 $\downarrow$} &
  \textbf{AUROC $\uparrow$} \\ \midrule
MSP \cite{MaxConfidenceScore}           & 79.04   & 77.38   & 80.94   & 75.87   &                      & 73.13    & 81.91    & 70.99   & 83.45   \\
ODIN \cite{ODIN}                        & 58.92   & 76.61   & 62.85   & 74.40   &                      & 63.14    & 82.59    & 59.82   & 82.20   \\
Energy score \cite{liu2020energy}       & 54.97   & 79.60   & 60.06   & 77.48   &                      & 58.69    & 82.98    & 56.89   & 83.69   \\
Gram matrices \cite{Gram}               & 77.55   & 59.38   & 60.93   & 74.93   &                      & 67.42    & 77.62    & 62.75   & 79.88   \\
Generalized ODIN \cite{baseline_G-ODIN} & 50.17   & 87.18   & 57.27   & 85.22   &                      & 70.28    & 79.23    & 59.57   & 83.12   \\
Mahalanobis \cite{ma_distance}          & 60.16   & 86.88   & 57.66   & 84.92   &                      & 96.27    & 57.42    & 96.46   & 59.25   \\
Vim \cite{wang2022vim}                  & 53.80   & 86.49   & 54.58   & 87.17   &                      & 88.40    & 68.73    & 83.47   & 71.94   \\
KNN \cite{sun2022out_knn}               & 44.50   & 88.37   & 47.28   & 87.45   &                      & 55.73    & 85.08    & 54.50   & 86.07   \\
CSI \cite{tack2020csi}                  & 37.06   & 87.99   & 47.10   & 84.09   &                      & 57.41    & 82.95    & 59.91   & 81.83   \\
MCM \cite{MCM}                  & 92.22   & 57.05   & 95.56   & 55.82   &                      & 71.52    & 81.45    & 62.47   & 83.15   \\
SIREN \cite{Siren}                      & 37.19   & 87.87   & 39.54   & 88.37   &                      & 49.12    & 87.21    & 54.23   & 86.89   \\
VOS \cite{VOS}                          & 35.61   & 88.46   & 44.13   & 86.92   &                      & 50.79    & 85.42    & 47.29   & 88.35   \\
TIB \cite{wu2023tib}                    & 24.00   & 92.54   & 36.85   & 88.47   &                      & 47.19    & 88.09    & 41.55   & 90.36   \\
SAFE \cite{wilson2023safe}              & 13.98   & 95.97   & 21.69   & 93.91   &                      & 17.69    & 94.38    & 36.32   & 87.03   \\ \midrule
\textbf{VisTa (Ours)} &
  \textbf{8.68} &
  \textbf{97.76} &
  \textbf{15.27} &
  \textbf{93.92} &
   &
  \textbf{9.49} &
  \textbf{97.91} &
  \textbf{25.74} &
  \textbf{94.47} \\ \bottomrule
\end{tabular}
\vspace{-1em}
\end{table*}

\subsection{Contextual Visual Prompt}
Our approach to object-level OOD detection begins by introducing carefully designed visual prompts. These prompts are layered on top of a fixed cropping operation that isolates the object of interest. The cropping step is consistently applied to all inputs, providing a uniform basis upon which additional visual prompts are added. The visual prompts are crucial for enhancing the model's ability to retain and leverage essential context, addressing the loss of such information when applying CLIP to object-level tasks. Each visual prompt emphasizes different aspects of the object and its surroundings, ultimately enriching the resulting visual features.

For example, the blur outside prompt focuses on the object by blurring the background, ensuring that the model concentrates on the object itself. The blur inside prompt blurs the details within the object while keeping the background clear, subtly highlighting the object's immediate environment. Additionally, drawing a distinctive color (e.g., red) around the bounding box creates a visual boundary that explicitly defines the object, enhancing the model's ability to differentiate between the object and its surrounding context.

Given an image $x$ and its bounding box $b$, each visual prompt $\phi_V^p(\cdot)$ generates specific visual features via the CLIP image encoder $\mathcal{I}(\cdot)$, denoted as:
\begin{equation}
    z_p=\mathcal{I}(\phi_V^p(x,b))    
    \label{equ:z_p}
\end{equation}

This formulation captures various aspects of the object's context and appearance. Then, these features are combined through element-wise addition to form a comprehensive visual representation:
\begin{equation}
    Z = {\sum_{p=1}^n z_p}/{\big \| \sum_{p=1}^n z_p \big \| }   
\end{equation}
where $n$ is the number of visual prompts and $\| \cdot \|$ denotes L2-norm. The integrated representation offers a holistic view of the object, encompassing fine-grained details and overall structure. 

\subsection{Text-Augmented ID Space Construction}
Given the complexity and richness introduced by the visual prompts, a simple textual description like “\textit{a photo of \{label\}}” is insufficient to represent the diversity of the ID space fully. To address this limitation, we propose a text-augmented approach to constructing the ID embedding space $\mathcal{C}$.

This approach enhances CLIP's text encoder by introducing text prompts aligned with the visual prompts applied earlier. For instance, to complement the blur outside visual prompt, we use a text prompt like “\textit{A photo of \{label\} with a blurred background}” ensuring that the textual embedding accurately reflects the visual emphasis on the object within its context. Similarly, the blur inside prompt is paired with a text prompt such as “\textit{A photo of \{label\} with blurred details}” to capture the nuanced focus on the object's environment.

By aligning these text prompts with the visual cues, we enrich the ID space $\mathcal{C}$ with a more expressive and contextually relevant set of embeddings. Specifically, $\mathcal{C}$ is constructed as:
\begin{gather}
    t_i^p=\phi_T^p(y_i),\\
    \mathcal{L}_i={\sum_{p=1}^n \mathcal{T}(t_i^p)}/{\big \| \sum_{p=1}^n \mathcal{T}(t_i^p) \big \|},\\
    \mathcal{C}=\{\mathcal{L}_i, i \in \{1,2,...,K\}\}
\end{gather}
where $\phi_T^p(\cdot)$ is the $p$-th prompt operation, $t_i^p$ is the $p$-th text prompt of label $y_i$, $\mathcal{T}(\cdot)$ is the CLIP text encoder, $n$ is the number of visual prompts, $\mathcal{L}_i$ is the augmented text embedding of label $y_i$ and $K$ is the number of ID classes. This enhanced ID space becomes crucial for accurately distinguishing between ID and OOD samples, allowing the model to understand better and interpret the contextual relationships inherent in the data.

\subsection{Uncertainty Estimation}

With the visual representation $Z$ and the enriched ID space $\mathcal{C}$ constructed, we compute the similarity score between them: $ S_{i} = \frac{Z\cdot \mathcal{L}_i}{\left \| Z \right \| \cdot \left \| \mathcal{L}_i  \right \| } $. This similarity represents the degree of matching between the object and the ID label. Then, we compute the uncertainty score:
\begin{equation}
    \mathbb{E}[\sigma(x, b) \mid \mathcal{C}] = -\text{log} \sum_{i=1}^{K} e^{ \tau \cdot S_i}
\end{equation}
where $\tau$ is a temperature parameter. The final uncertainty score is then used to differentiate between ID and OOD samples as in \eqref{equ:case}.

\begin{table}[t]
\centering
\caption{Impact of our contextual visual prompt (VP) and text-augmented ID space construction (TA).}
\label{tab:ab_study_vp_tas}
\begin{tabular}{ccccc}
\toprule
\multirow{2}{*}{\textbf{ID dataset}} & \multicolumn{2}{c}{\textbf{Module}} & \multicolumn{2}{c}{\textbf{OpenImages/MSCOCO}}          \\ \cmidrule(l){2-5} 
                                     & \textbf{TA}      & \textbf{VP}      & \textbf{FPR95 $\downarrow$} & \textbf{AUROC $\uparrow$} \\ \midrule
\multirow{4}{*}{\textbf{BDD-100k}}     & -          & -          & 64.12 / 69.16         & 72.01 / 71.83          \\
                              & \checkmark & -          & 60.12 / 63.18         & 75.29 / 74.37          \\
                              & -          & \checkmark & 23.42 / 30.96         & 91.90 / 90.37          \\
                              & \checkmark & \checkmark & \textbf{8.68 / 15.27} & \textbf{97.76 / 93.92} \\ \midrule
\multirow{4}{*}{\textbf{VOC}} & -          & -          & 46.80 / 55.54         & 88.23 / 76.59          \\
                              & \checkmark & -          & 38.97 / 51.37         & 91.78 / 88.92          \\
                              & -          & \checkmark & 20.76 / 30.31         & 92.01 / 91.17          \\
                              & \checkmark & \checkmark & \textbf{9.49 / 25.74} & \textbf{97.91 / 94.47} \\ \bottomrule
\end{tabular}
\vspace{-1em}
\end{table}

\section{Experiments}

\subsection{Experimental setup}

\textbf{Datasets.} We perform object detection tasks using the predefined ID/OOD splits outlined in\cite{VOS}. The two ID datasets are derived from the widely used PASCAL-VOC\cite{VOC} and BDD-100K\cite{yu2020bdd100k} datasets. For the OOD datasets, we provide subsets of the MS-COCO \cite{mscoco} and OpenImages \cite{openimages} datasets, ensuring that classes present in the custom ID datasets are excluded.

\textbf{Evaluation metrics.} Following \cite{VOS}, we adopt two key metrics: \textbf{FPR95} and \textbf{AUROC}. VisTa serves as a supplementary component to an already trained object detection network and does not influence the base model's performance regarding the mean average precision (mAP) metric; therefore, we do not include mAP in our reporting as done in \cite{VOS}.

\begin{table}[ht]
\centering
\caption{Impact of different CLIP backbone. (CROP ONLY)}
\label{tab:vb_clip_backbone}
\begin{tabular}{cccccc}
\toprule
\textbf{Backbone}    & \textbf{RN50}  & \textbf{RN101} & \textbf{VIT-B/32} & \textbf{VIT-B/16} & \textbf{VIT-L/14} \\ \midrule
\textbf{FPR95}       & 52.34 & 50.23 & 48.31    & 46.80    & 46.59    \\ \midrule
\textbf{AUROC}       & 84.97 & 85.79 & 87.19    & 88.23    & 88.37    \\ \midrule
\textbf{Runtime(ms)} & 86.7  & 154.2 & 91.3     & 223.7    & 866.9    \\ \bottomrule
\end{tabular}
\vspace{-1em}
\end{table}

\begin{figure}[ht]
    \centering
    \includegraphics[width=\linewidth]{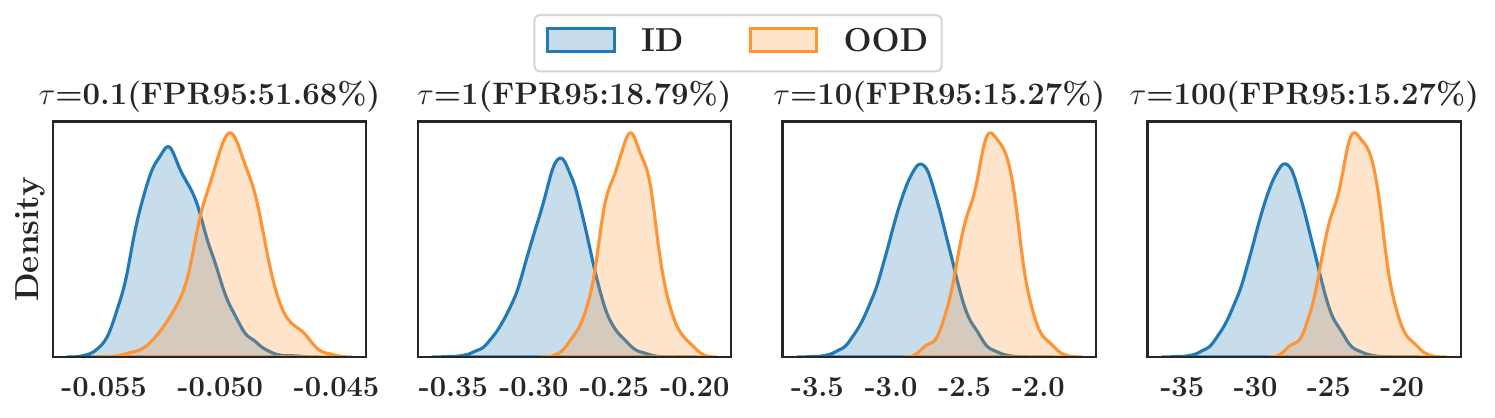}
    \caption{Impact of different temperature parameter $\tau$.}
    \label{fig:ab_tau}
    \vspace{-1em}
\end{figure}

\textbf{Implementation details.} We implement the Faster-RCNN detector with ResNet-50 using the Detectron2 library \cite{wu2019detectron2} and employ CLIP (ViT-B/16 \cite{2021An}) as the VLM. For the visual prompts, we adopt crop, blur inside, blur outside, and box. We configured the Gaussian blur with a standard deviation of 2 and chose red for the box color.

\subsection{Main Results}

\setlength{\tabcolsep}{5pt}

As shown in Table \ref{tab:main_tab}, our proposed zero-shot CLIP-based approach, VisTa, demonstrates advantages over previous methods. Notably, on the autonomous driving dataset BDD-100K, VisTa greatly enhances the identification of OOD objects. In tests on the OOD dataset OpenImages, VisTa achieves an FPR95 of \textbf{8.68\%}, a \textbf{5.30\%} reduction compared to the previously best-performing method, SAFE. On the OOD dataset MSCOCO, VisTa achieves an FPR95 of \textbf{15.27\%}, improving by \textbf{6.42\%} compared to SAFE. When VOC serves as the ID dataset and OpenImages as the OOD dataset, VisTa performs the best, achieving an FPR95 of \textbf{9.49\%}, which is an improvement of \textbf{8.20\%} compared to SAFE. On the OOD dataset MSCOCO, VisTa achieves an FPR95 of \textbf{25.74\%}, improving by \textbf{10.58\%} compared to SAFE. These results highlight VisTa’s robustness and strong OOD detection capabilities in both zero-shot and non-zero-shot scenarios.

\subsection{Analysis}

\textbf{Impact of visual prompt and text-augmented ID space.} Table \ref{tab:ab_study_vp_tas} illustrates the impact of visual prompt and text-augmented ID space construction on OOD detection performance. The results show that while visual prompts alone significantly improve detection accuracy, adding the text-augmented ID space without corresponding visual prompts offers minimal performance gains. This minimal gain is because the textual enhancements are designed to complement the visual cues, and without the visual prompts, the model cannot fully leverage the added semantic layer. Visual prompts are essential for preserving contextual information, which is crucial for distinguishing objects at the local level. Combined with the text-augmented ID space, which adds a complementary semantic layer, the two components synergistically improve feature representation, directly supporting our goal of leveraging visual and textual cues for robust OOD detection.


\textbf{Impact of different CLIP backbone.} We evaluate ResNet50, ResNet101, ViT-B/32, ViT-B/16, and ViT-L/14 using VOC (ID) and OpenImages (OOD) datasets with batch size 8. ResNet50 is fastest but provides weaker features, while ResNet101 improves performance at higher cost. ViT-B/32 balances performance and efficiency, and ViT-B/16 further enhances performance with increased computation. ViT-L/14 achieves the best results but with marginal gains over ViT-B/16 and significantly higher costs. ViT-B/16 is optimal, balancing performance and efficiency, while ViT-L/14 highlights the trade-off between performance and cost.

\begin{figure}[htbp]
    \centering
    \begin{minipage}[c]{\linewidth}
        \centering
        \includegraphics[width=\textwidth]{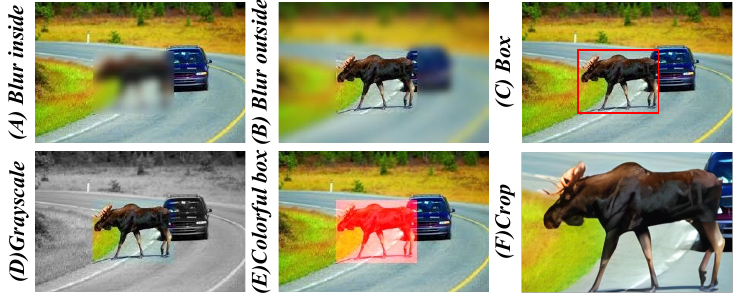}
        \centerline{\scriptsize (a) Illustration of different visual prompts.}
    \end{minipage} \\
    \vspace{1em}
    \begin{minipage}[c]{\linewidth}
        \centering
        \includegraphics[width=\textwidth]{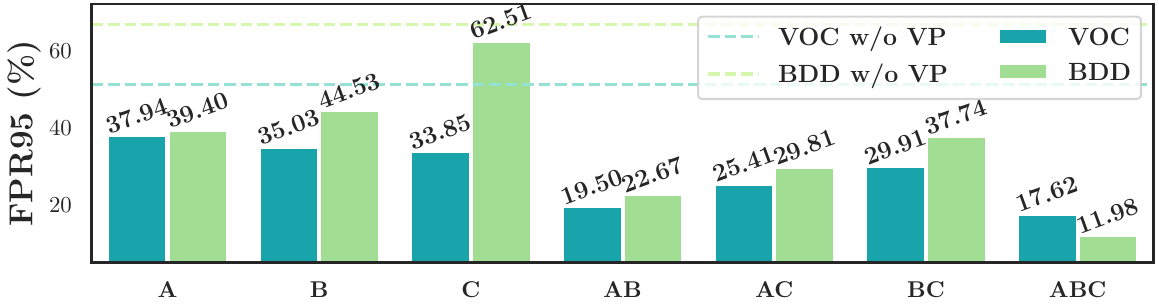}
        \centerline{\scriptsize (b) Qualitative results of different visual prompts.}
    \end{minipage}
    \vspace{0.5em}
    \caption{Impact of different visual prompts. We report average results. Text-augmented ID space is constructed with corresponding visual prompts.}
    \label{fig:vp_ablation}
    \vspace{-1em}
\end{figure}

\textbf{Impact of temperature parameter $\tau$.} Using BDD\cite{yu2020bdd100k} as ID and MSCOCO\cite{mscoco} as OOD datasets, we study the effect of $\tau$ (0.1, 1, 10, 100) on ID-OOD separation. Fig. \ref{fig:ab_tau} shows that moderately increasing $\tau$ improves separation, but beyond a threshold, further increases provide no benefit. This indicates $\tau$ is most effective within a moderate range, optimizing similarity computation without over-smoothing.

\textbf{Impact of different visual prompts.} In our ablation study, we evaluate three specific visual prompting techniques: (A) blur inside, (B) blur outside, and (C) box, analyzed alongside a fixed cropping operation to isolate the object of interest. Fig. \ref{fig:vp_ablation}(a) illustrates the visual effects of these prompts, along with (D) grayscale, (E) colorful box, and the standard (F) crop operation. Blur inside preserves the surrounding context, while blur outside emphasizes the object by softening its background. The colorful bounding box enhances visual salience. Although we explore grayscale and colorful boxes, their minimal impact on feature extraction and occasional performance degradation leads us to exclude their data from the main bar chart.



The bar chart in Fig. \ref{fig:vp_ablation}(b) quantifies the primary visual prompts' impact on OOD detection accuracy. Blur inside shows the most improvement, likely due to VLMs' pre-training on "Bokeh"-style datasets. Blur outside also performs well, while the colorful box shows modest gains, reflecting its limited influence compared to blur methods.



\section{Conclusion}

We present an innovative zero-shot method for object-level OOD detection that combines visual prompts with text-augmented ID space construction. Our method enhances CLIP's ability to preserve crucial contextual information and enriches the ID embedding space with aligned textual cues. This integration enables strong performance across multiple benchmarks in a zero-shot setting. Experimental results show that our approach consistently surpasses prior methods. Furthermore, unlike existing methods that require retraining the detector or adding new networks, our approach works directly with pre-trained models, maintaining high ID accuracy while significantly improving OOD detection without additional training.




\bibliographystyle{ieeetr}
\bibliography{main}

\end{document}